\documentclass[letterpaper, 10 pt, conference]{ieeeconf}
\IEEEoverridecommandlockouts
\usepackage{graphics} % for pdf, bitmapped graphics files
\usepackage{float}
\usepackage{epsfig} % for postscript graphics files
\usepackage{mathptmx} % assumes new font selection scheme installed
\usepackage{times} % assumes new font selection scheme installed
\usepackage{amsmath} % assumes amsmath package installed
\usepackage{amssymb} % assumes amsmath package installed
\usepackage[ruled]{algorithm2e}
\usepackage{color}
\usepackage[colorlinks,linkcolor=blue,citecolor=red,anchorcolor=green,backref]{hyperref}

\title
{\LARGE \bf
RMMDet: Road-Side Multitype and Multigroup Sensor Detection System for Autonomous Driving
}

\author{Xiuyu Yang$^{1*}$, Zhuangyan Zhang$^{2*}$, Haikuo Du$^{3}$, Sui Yang$^{2}$, Fengping Sun$^{1}$, Yanbo Liu$^{4\dag}$, \\Ling Pei$^{4}$, Wenchao Xu$^{5}$, Weiqi Sun$^{6}$, Zhengyu Li$^{7}$
\thanks{This work was supported in part by Cooperative education program of the Ministry of Education 202101001039 and in part by the National Natural Science Foundation of China under Grant Nos. 61873163. *These authors contributed equally to this work.
}
\thanks{$^{1}$Xiuyu Yang and Fengping Sun are with the Department of Electronic Engineering of SEIEE, Shanghai Jiao Tong University. {\tt\small \{gzzyyxy, sunfengping\}@sjtu.edu.cn}}
\thanks{$^{2}$Zhuangyan Zhang and Sui Yang are with the Department of Electrical Engineering of SEIEE, Shanghai Jiao Tong University. {\tt\small \{zhangzhuangyan, suiyang\}@sjtu.edu.cn}}
\thanks{$^{3}$Haikuo Du is with the Department of Automation of SEIEE, Shanghai Jiao Tong University. {\tt\small duhaikuo1013@sjtu.edu.cn}}
\thanks{$^{\dag}$\,is the corresponding author. $^{4}$Yanbo Liu and Ling Pei are with Teaching development and Student Innovation Center of SEIEE, Shanghai Jiao Tong University. {\tt\small \{liuyanbo1205, ling.pei\}@sjtu.edu.cn}}
\thanks{$^{6}$Wenchao Xu is with Shanghai Key Laboratory of Multidimensional Information Processing, East China Normal University. {\tt\small wchxu@ce.ecnu.edu.cn}}
\thanks{$^{7}$Weiqi Sun is with Xin Dong Interactive Entertainment Company Ltd, Shanghai. {\tt\small sunweiqi7777@gmail.com}}
\thanks{$^{7}$Zhengyu Li is with Summer Son Smart Technology Company Ltd, Shanghai. {\tt\small zhengyuli@gmail.com}}
}

\begin{document}
\setlength{\abovecaptionskip}{0.cm}
\maketitle
\thispagestyle{empty}
\pagestyle{empty}

\begin{abstract}
Autonomous driving has now made great strides thanks to artificial intelligence, and numerous advanced methods have been proposed for vehicle end target detection, including single sensor or multi-sensor detection methods. However, the complexity and diversity of real traffic situations necessitate an examination of how to use these methods in real road conditions. In this paper, we propose RMMDet, a road-side multitype and multigroup sensor detection system for autonomous driving. We use a ROS-based virtual environment to simulate real-world conditions, in particular the physical and functional construction of the sensors. Then we implement muti-type sensor detection and multi-group sensors fusion in this environment, including camera-radar and camera-lidar detection based on result-level fusion. We produce local datasets and real sand table field, and conduct various experiments. Furthermore, we link a multi-agent collaborative scheduling system to the fusion detection system. Hence, the whole roadside detection system is formed by roadside perception, fusion detection, and scheduling planning. Through the experiments, it can be seen that RMMDet system we built plays an important role in vehicle-road collaboration and its optimization. The code and supplementary materials can be found at: \href{https://github.com/OrangeSodahub/RMMDet}{https://github.com/OrangeSodahub/RMMDet}

\end{abstract}

\section{INTRODUCTION}

    \begin{figure}[thpb]
        \centering
        \includegraphics[scale=0.38]{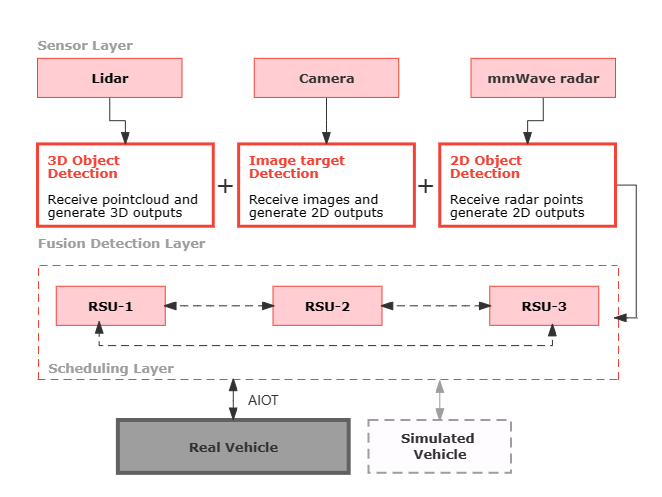}
        \caption{Overall Structure of RMMDet}
        \label{Overall Structure of RMMDet}
    \end{figure}

Now for autonomous driving, the self-control and decision are obtained based on the collection of massive traffic states and intensive information processing. Note that the spatial-temporal characteristics of the traffic states and the constrained perception range of an individual vehicle seriously undermine the effectiveness of the state collection, and sensors have their weakness. However, roadside sensing provides extended coverage and more information compared with independent vehicle sensors, and now many multi-sensor fusion strategies are proposed to achieve complementary advantages. In \cite{c1}, the author proposed to combine color image information and point cloud geometry information by using MobelNet V3 as the backbone and migrated RPN layer, this work basically introduces two modal data fusion to us. As for the camera-radar information use, RRPN \cite{c2} mapped radar detection results to the image plane for anchor frames generation, so it also brings large computation reduction. CenterFusion \cite{c3} proposed the frustum association mechanism to realize the radar and image detection fusion. In contrast, more work has been done on lidar since its strong characteristics and ability to carry more severe and complex detection tasks. Generally it could be divided into: lidar-only methods \cite{c4}-\cite{c6}, image-only methods \cite{c7} \cite{c8} and lidar-RGB image methods \cite{c9}-\cite{c12}. While less work focused on camera-radar-lidar fusion detection \cite{c13} \cite{c14} or radar-lidar fusion \cite{c15}. Although these fusion work continued the SOTA performance on detecting or tracking objects, unfortunately, could suffer from the spatial-temporal misalignment inherent in point fusion \cite{c12}, or inaccurate correspondence among dense or sparse feature vectors for feature-level fusion \cite{c11}, as CLOC \cite{c19} and LIF-Seg \cite{c20} proved. For roadside sensing and detection, previous work \cite{c16} \cite{c17} \cite{c18} built a system more useful to intelligence transport relying on camera-radar or vision only, despite the demonstrated success, there exists limitations on its single application scenario, such as the inner city straight road \cite{c17} for traffic activities surveillance or the suburban highway \cite{c18}.

As for an real roadside detection system for auto-driving, we identify the key points to (i) Efficiency (i.e. high speed and low memory)---undertake the task of real time data acquisition, transmission and  inference, and even scheduling. (ii) Robustness---ensure that system will not fail under abnormal conditions. Considering the interdependence and computation in data-level fusion and feature-level fusion detection algorithms \cite{c9}-\cite{c15}, we chose to establish a result-level fusion mechanism. (iii) Modularity---use any pair of 2D and 3D detectors as modules, which will be easily for system renovation and experiments. And (iv) Virtual counterpart---use a simple yet functional simulation environment with strong correlation with the real conditions for the maximum plasticity and convenience. In this paper, we present \textbf{RMMDet}, a roadside multi-type and multi-group sensor detection system for autonomous driving. We use ROS \cite{c21} to model all the elements, including the environment, vehicles and sensors. We adopt PointPillars \cite{c5} and YOLOv4 \cite{c22} as our independent 3D and 2D detection algorithms. We use the method adapted from EKF \cite{c23} and RRPN \cite{c2} to execute the radar-side prediction and radar-camera detection, and NMS-like mechanism to help with the lidar-camera detection.

Additionally, we add a multi-agent collaborative scheduling system, so as to expand our work to a more comprehensive level. Some researches \cite{c24} on vehicle-road cooperation introduced multi-mode information fusion mechanism, which expands the range of traffic information exchange. Since all the work mentioned above have carried out hierarchical research from various aspects, here we try to consider them as a whole and perform experiments.

This paper contributes:

\begin{enumerate}
    \item A comprehensive simulation environment based on ROS integrating multi-type sensors physically and functionally which reaches mentioned efficiency, robustness and modularity.
    \item An end-to-end multi-type and multi-group sensor detection system initially running on the general algorithms for autonomous vehicles, which supports the interface between algorithmic reasoning and data streams and any other algorithm tests.
    \item Virtual and actual experiments on this detection system based on downstream module (i.e. multi-agent collaborative scheduling system) to evaluate the whole real-time work and it validates our idea of integrating perception, detection and decision.
\end{enumerate}

\section{SIMULATION ENVIRONMENT}

ROS \cite{c21} is used to construct a 1:18 scale simulation model in this paper. We first build models in SolidWorks based on accurate scaling and component connection and import them into ROS through URDF files. Completing the sensor function configuration, finally, it can achieve the information acquisition and transmission of various sensors.

\subsection{Traffic Environment}

We design a rectangular field of $8m\times 10m$ (width and length) which takes the actual situation into full consideration, including viaduct, ring road, intersection, straight road, turning road, and other common elements.

    \begin{figure}[thpb]
        \centering
        \includegraphics[scale=1.0]{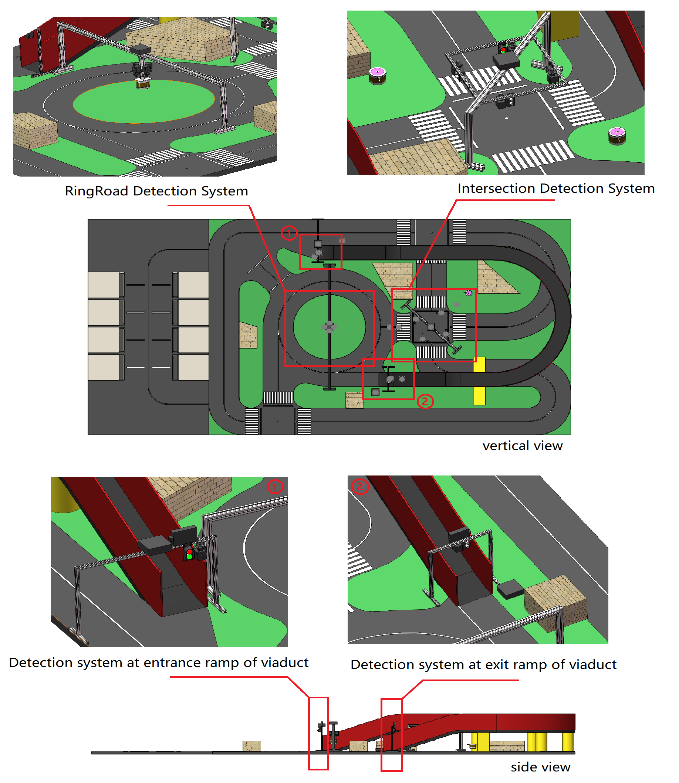}
        \caption{BEV and details of environment}
        \label{BEV and details of environment}
    \end{figure}

As shown in Figure \ref{BEV and details of environment}, among them, the detection system is deployed in the circular section, intersection, and viaduct ramps. Considering the traffic flow around the ring road and the intersection is relatively complex and heavy, we set four monocular cameras positioned 90° apart in four directions, and one 32-line lidar  centrally suspended for the circular section, two 16-line lidars diagonally located for the intersection, where every four cameras could cover 360° with additional overlaps.

At the viaduct sections, vehicles are usually aligned with long lines, not particularly clumpy or clustered distribution, and the directions are not irregularly staggered at multiple angles, where millimeter wave radar plays a greater role in measuring the speed, position, and other information of the target accurately. Therefore, we set the camera-radar detection system, which is also suitable for others with similar characteristics such as the highways \cite{c18}.

In addition, we also set up the driving environment of the real vehicle (See Experiments Sec.\ref{Experiments}) to validate our scheme in real world.

\subsection{Sensors: camera, radar and lidar} \label{sensors}

Bodies of sensors are modeled according to their life-sizes and the scale. Then they are functionally configured to capture and transmit massive data, specifically using the camera plug-in in ROS, Robosense lidar plug-in\footnote{Robosense lidar plug-in: \href{https://github.com/RoboSense-LiDAR/rslidar_sdk}{https://github.com/RoboSense-LiDAR/rslidar-sdk}} and millimeter wave radar plug-in\footnote{Millimeter wave radar plug-in: \href{https://github.com/OrangeSodahub/Delphi-ESR-SSR-gazebo-plugin}{https://github.com/OrangeSodahub/Delphi-ESR-SSR-gazebo-plugin}} for the image acquisition, point cloud generation and radar points acquisition. Table \ref{Attributes of sensors and data} shows the attributes of sensors and their data stream. "FPS" and "Shape" are both in the simulation environment running on CPU intel i7 and GPU RTX3080.
\begin{table}[h!]
    \centering
    \caption{Attributes of sensors and data}
    \label{Attributes of sensors and data}
    \begin{tabular}{ c|c c c } 
        \hline
        \rule{0pt}{7pt}
        Sensor & Type & FPS & Shape \\
        \hline
        \rule{0pt}{7pt}
        Camera & Deepracer & $\sim$180 & 1920$\times$1080 /pixels \\
        16-line Lidar & RoboSense & $\sim$50 & $\sim$2000 /points \\
        32-line Lidar & RoboSense & $\sim$50 & $\sim$4000 /points \\
        Radar & ARS\_408\_21 & $\sim$200 & 10$\sim$20 /points \\
        \hline
    \end{tabular}
\end{table}

\section{OVERALL FRAMEWORK}
Figure \ref{Overall Structure of RMMDet} illustrates the main framework of the whole system with three parts: sensor layer, fusion perception layer, and decision planning layer. The sensor layer contains physical devices of three types of sensors which have been constructed in \textit{Sensors} (Sec. \ref{sensors}).

The fusion detection layer contains multi-type sensor target detection and multi-group sensor fusion detection. After preprocessing the raw data obtained from the sensors, feed them into the separate branches of image detection based on cameras, 2D point cloud detection based on the radars, and 3D point cloud detection based on lidars. This operation could be performed in the respective mobile edge computing platform (MEC). Then different fusion strategies enhance the detection and output final results, including camera-radar group and camera-lidar group. Additionally, the unique information participated in fusion, such as target velocity detected by millimeter wave radar, etc, will also be hold in outputs.

Multi-agent Collaborative system, a lightweight module for downstream tasks, is realized in the Decision planning layer.Based on different decision-making agents divided according to the traffic function or spatial-level characteristics, each body detects within its area respectively and carries out the information transmission among the vehicles and bodies.

\section{RMMDet: Road-Side Multitype and Multigroup Sensor Detection System}

\subsection{Coordinate System} \label{Coordinate System}

\begin{figure}[thpb]
        \centering
        \includegraphics[scale=1]{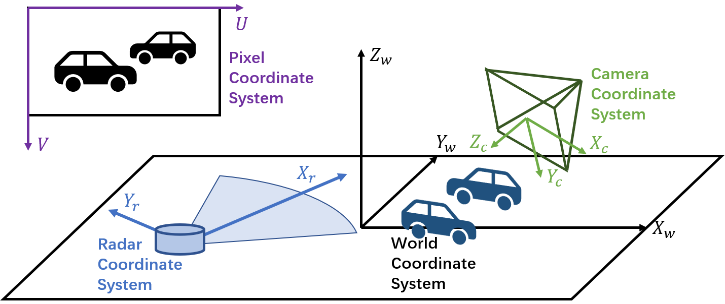}
        \caption{Sketch of four coordinate systems}
        \label{Sketch of four coordinate systems}
    \end{figure}

Four coordinate systems are depicted in Fig.\ref{Sketch of four coordinate systems}: camera(pixel), radar, lidar, world coordinates.The subsequent steps involve the transformations of these four coordinate systems to ensure the unity of spatial representation of different data, which can be summarized as following equations:

\begin{equation} \label{eq1}
    \textbf{C}_w^h
    =
    \left[
    \begin{array}{cc}
        \textbf{R} & \textbf{T} \\
        \textbf{0} & 1
    \end{array}
    \right]
    \textbf{C}_{c,r,l}^h
\end{equation}

\begin{equation} \label{eq2}
    Z_c
    \cdot
    \textbf{C}_{u,v}^h
    =
    \textbf{PK}
    \cdot
    \textbf{C}_c^h
    =
    \textbf{PKE}
    \cdot
    \textbf{C}_w^h
\end{equation}

In eq.\ref{eq1}, three types of subscripts of homogeneous coordinates \textbf{$C^h_{\{\cdot\}}$} denote the pixel frame (i.e. \textbf{$C_{u,v}^h$}), world frame (i.e. \textbf{$C_w^h$}) and reference frame (i.e. \textbf{$C_{c,r,l}^h$}) respectively, where the reference represents the sensors. Other forms of this formula could project the target from any frame to the other one. And Eq.\ref{eq2} enables the projection from world or camera coordinates to the 2-dimensional pixel coordinates, where $\textbf{K}$ and $\textbf{E}$ represents the intrinsic and extrinsic camera parameter matrix. Matrix $\textbf{P}$ transforms the unit from meter to pixel and produces the pixel coordinates \textbf{$C_{u,v}^h$}.

\subsection{Multi-type Sensor Detection}

As mentioned before, an ideal road-side detection system must be efficient and robust to perform high-load real-time detection tasks. Most of detectors \cite{c9}-\cite{c15} uses the features after fusion as the input of prediction head. Unfortunately, this results in strong correlation and interdependence among different modal data. Once any of them is contaminated or fails, others associated will not contribute to the detection system. Thus we turn to result-level fusion mechanism and build independent multi-type sensor detection in advance.

    \begin{figure}[thpb]
        \centering
        \includegraphics[scale=1]{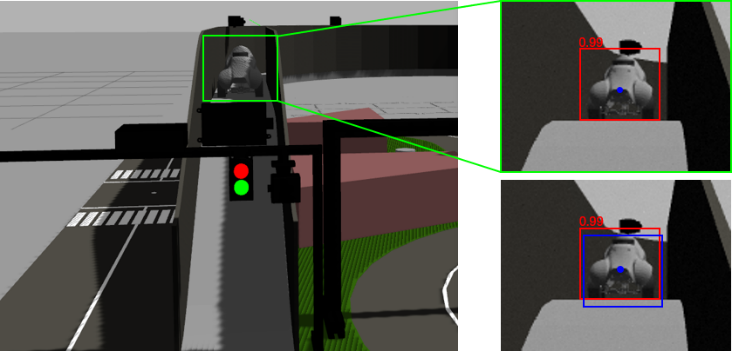}
        \caption{Radar ROI generation:  the left is the original image, and the right shows the POI (blue point) and the expanded ROI (blue box) of the target.}
        \label{Radar ROI generation}
    \end{figure}

    \begin{figure*}
        \includegraphics[width=\linewidth]{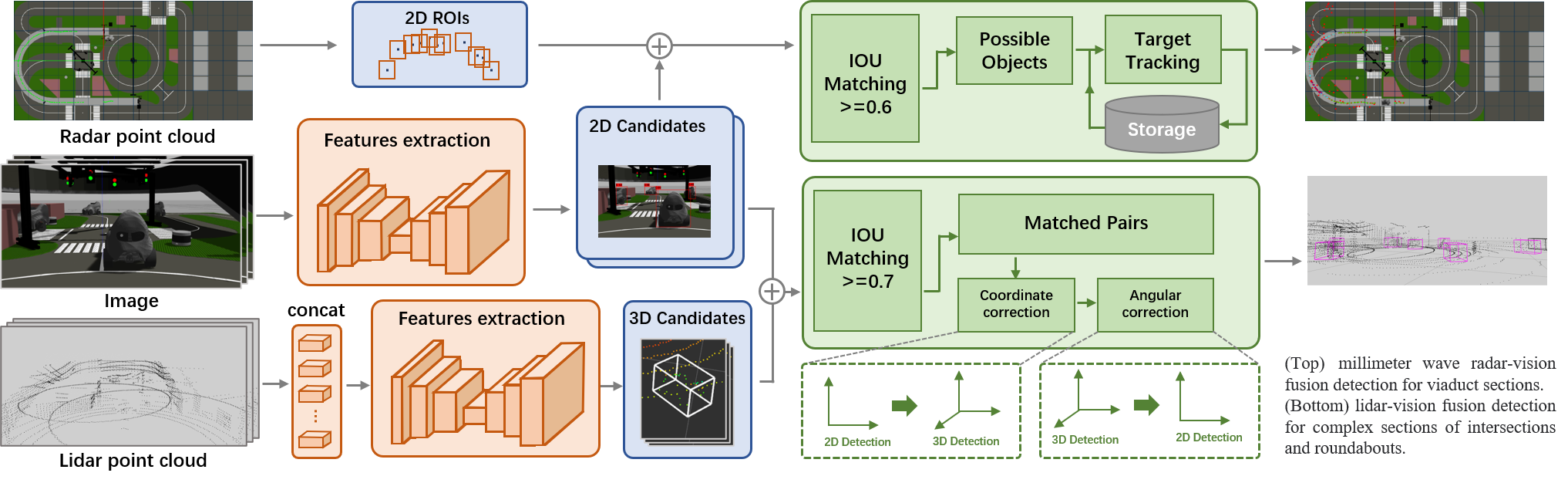}
        \caption{Fusion Detection Layer-CRLFnet: Input the raw data (radar and lidar point cloud, image) to separate the detection model to generate ROIs. And through the fusion part (green boxes), final detection results with high confidence are output.}
        \label{Fusion Detection Layer}
    \end{figure*}

\textbf{Radar.} Inspired by RRPN \cite{c2} that generates anchors based on projected radar points on images, and CenterFusion \cite{c3} that expands the radar points to pillars to eliminate the variance of data in z-dimension, after the projection from radar coordinates to camera coordinates and producing the points of interests (POIs) on image, we expand the discrete points to the region of interests (ROIs) based on the known target sizes and the distance compensation. We apply the scaling factor to the predefined sizes to complete the distance compensation:

\begin{equation} \label{eq3}
    S_i = \frac{\alpha}{d_i} + \beta
\end{equation}

In eq.\ref{eq3}, $S_i$ is the scaling factor, $d_i$ is the perpendicular distance along the normal detection plane to the target, $\alpha$ and $\beta$ are two parameters determined by the size of the vehicle. The process of radar ROIs generation is shown in Figure \ref{Radar ROI generation}.

\textbf{Camera and Lidar.} YOLOv4 \cite{c22} and PointPillars \cite{c5} are our independent 2D detection and 3D point cloud detection tools. According to the report of PointPillars \cite{c5}, it has excellent detection accuracy especially low latency which enables the real-time applications. And YOLOv4 achieved 43.5\% AP (65.7\% AP50) for the MS COCO dataset \cite{c25} at a real-time speed of $\sim$65 FPS on Tesla V100. Other advanced algorithms could also be added to the system.

Point cloud data from different lidars are merged first to reach the unified spatial representations and enhance the perception information, that is, apply the coordinates transformation based on eq.\ref{eq1} and concatenation. And the whole size of points data in our simulation environment is up to $6000\sim 7000$, of which invalid points account for a large proportion. To greatly improve the efficiency, following M. Himmelsbach et al. \cite{c26}, we assign the Gaussian filter and RANSAC to filter the noise, outliers and ground points, which brings reduction by around 40\%.

\textbf{Data synchronization.} Though multiple detectors work independently, different data stream should be synchronized to ensure the valid results. Specifically, \textit{Coordinate system} (Sec. \ref{Coordinate System}) already achieves the spatial synchronization. And we align the timestamps of data steams released by ROS to get the temporal synchronization.

\subsection{Multi-group Sensor Fusion Detection}
\textbf{Millimeter-wave Radar and Camera Fusion.} The fusion layer takes in the ROIs from the radars and the cameras and returns the world coordinates of the detected objects. It includes two steps, IOU matching and target tracking, as shown in Fig.\ref{Fusion Detection Layer}.

The IOU matching step identifies potential objects by the ROIs from radars and cameras, whose correlation is evaluated by the intersection over union (IOU). As shown in Algorithm \ref{Radar-Camera IOU Matching}, the program traverses the ROIs, calculates the IOU matrix, separates the raw detection results into pairs (potential objects) by the K-M algorithm, sorts them by IOU and discards those have low correlation (IOU $<$ 0.6). The pixel coordinates of each pair are transformed to the world coordinates and then returned.

\begin{algorithm}
    \caption{Radar-Camera IOU Matching}
    \label{Radar-Camera IOU Matching}
    \LinesNumbered
    \KwIn{Radar ROIs and Camera ROIs}
    \KwOut{Concatenated Pixel Coordinates of Potential Objects}
    \For{each radar ROI and camera ROI}{
        calculate the IOU of radar ROI and camera ROI\;
    }
    matched pairs = K-M algorithm (IOU matrix)\;
    \For{each pair of matched pairs}{
        \eIf{IOU $<$ 0.6}{
            remove the pair from matched pairs\;
        }{
            concatenate the pixel coordinates\;
        }
    }
    \Return concatenated pixel coordinates of potential objects\;
\end{algorithm}

\begin{figure}[thpb]
        \centering
        \includegraphics[width=0.48\textwidth]{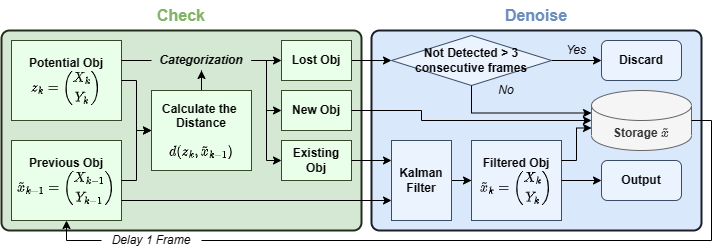}
        \caption{Process of target tracking. The variant $z$ is the coordinate of potential objects, $x$ is the coordinate of filtered objects.}
        \label{Process of target tracking}
    \end{figure}

The target tracking step checks the potential objects and denoises. As shown in Fig.\ref{Process of target tracking}, the program calculates the distance between previous detected objects and the potential objects and then categorizes the potential objects as new, lost, and existing ones. The new objects are not returned for stability, while the lost ones will be discarded if not being detected for over 3 consecutive frames. Two nearing objects in two frames are regarded as an existing object, whose coordinates are filtered by Kalman Filter and then returned.

The time update equations calculate the prior state estimate based on the results of the last frame.

\begin{equation} \label{eq4}
    \begin{cases}
        \tilde{x}_k^- &= \mathbf{A}\tilde{x}_{k-1}\\
        \mathbf{P}_k^- &= \mathbf{AP}_{k-1}\mathbf{A}^\text{T}+\mathbf{Q}
    \end{cases}
\end{equation}

The measurement update equations calculate the posterior state estimate based on the prior state estimate and the measurement. The Kalman gain $K$ is the ratio of predicted error to measurement error.

\begin{equation} \label{eq5}
    K = \mathbf{P}_k^ - {\mathbf{H}^\text{T}}{{\left( \mathbf{HP}_k^- \mathbf{H}^\text{T} + \mathbf{R} \right)}^{-1}}
\end{equation}
\begin{equation} \label{eq6}
    \begin{cases}
        \tilde x_k = \tilde x_k^-  + K\left( z_k - \mathbf{H} \tilde x_k^-  \right)\\
        \mathbf{P}_k = \left( \mathbf{I} - K\mathbf{H} \right) \mathbf{P}_k^- 
    \end{cases}
\end{equation}

As shown in Eq.\ref{eq4}-\ref{eq6}, $x_k$ is the real world coordinate of an existing object, $z_k$ is the observed one. $\mathbf{P}_k$ is the state covariance matrix, which we try to minimize. The transition matrix $\mathbf{A}$ and the observation matrix $\mathbf{H}$ are identity matrices in this article. The transition covariance $\mathbf{Q}$ and the observation covariance $\mathbf{R}$ are manually set.

After that, the world coordinates of new and existing objects are stored in the storage and wait to be loaded in the next frame.

\textbf{Lidar and Camera Fusion.} The fusion strategy is shown in Fig.\ref{Fusion Detection Layer}. The data obtained by the lidar and the camera are processed and then fed into separate detection branches. 

YOLOv4 \cite{c22} and PointPillars \cite{c5}, two conventional and popular algorithms, are used for different model object detection. We get two types of predicted boxes at the ends of the branches, $D^{\,C}\in R^{\,N\times 8}$ and $D^{\,L}\in R^{\,L\times M\times 4}$. To get the last single result, as shown in the Algorithm \ref{Lidar-Camera Box Matching}, boxes are concatenated and matched. Specifically, for each 3D detection box, there exists its multiple 2D detection boxes lie on different image frame, due to the overlap of range of cameras, and those with larger areas on the image plane are relatively more accurate, in this regard, we sort them based on the distances from the 3D prediction to the cameras. This operation only depends on the soft physical associations among sensors, which nearly has no influence on the accuracy. We mark the matched box of the first camera (geographically nearest) in the sorted camera list, and mask the matched boxes of others. By traversing the camera list and all of 2D boxes for each 3D box, we calculate the IOU to determine whether each 2D box is marked, masked or discarded, except the marked or masked ones in the former steps. When $N$ 3D detection boxes in the scene are traversed, all the 2D detection boxes are traversed $N$ times, hence we discard those who are still not matched.

\begin{algorithm}
    \caption{Lidar-Camera Box Matching}
    \label{Lidar-Camera Box Matching}
    \LinesNumbered
    \KwIn{Raw detection boxes $D^{\,C}$ and $D^{\,L}$}
    \KwOut{Merged detection boxes}
    matched boxes: $\Dot{D^{\,C}}, \Dot{D^{\,L}} = [\,], [\,]$\;
    \For{$d^l_{i,\ i=1,2,3...}$ in $D^{\,L}$}{
        calculate the distances array $A=\{||d^l_i-c_j||\}$\;
        generate the ordered list $C_i$ and $D^{\,C}_i$ sorted by $A$\;
        \For{$c_{ij,\ j=1,2,3...}$ in $C_i$}{
            map $d^l_i$ to the image plane of $c_{ij}$ and get $\tilde{d^l_i}$\;
            \For{$d^c_{ijk,\ k=1,2,3...}$ in $D^{\,C}_{ij}$}{
                skip if $d^c_{ijh}$ exist in $\Dot{D^{\,C}}$\;
                calculate IOU of $d^c_{ijk}$ and $\tilde{d^l_i}$ and form the array $V_{ij}$\;
            }
            get the max IOU from the list\;
            \If{$\underset{j}{max}\ V_{ij} \geqslant 0.7$}{
                get the corresponding 2d box $d^c_{ijh}$\;
                \eIf{$d^l_i$ not in $\Dot{D^{\,L}}$}{
                    add $d^l_i$ to $\Dot{D^{\,L}}$, and add $d^c_{ijh}$ to $\Dot{D^{\,C}}$\;
                }{
                    delete $d^c_{ijh}$ from $D^{\,C}$\;
                }
            }
        }
    }
    Discard unmatched 2d and 3d boxes\;
\end{algorithm}

Merging is after matching. By projecting the center of image box to 3D world given the height (z-axis) of the target, we get finalize the box center based on weighted coordinates. And adjust the azimuth angle of coordinates-fixed 3d box by adding or subtracting angle increments until the aspect ratio of the lidar-based 2d box on the image plane is approximately equal to the weighted mean ratio. Generally, this step does not require much computation.

    \begin{figure}[thpb]
        \centering
        \includegraphics[scale=0.45]{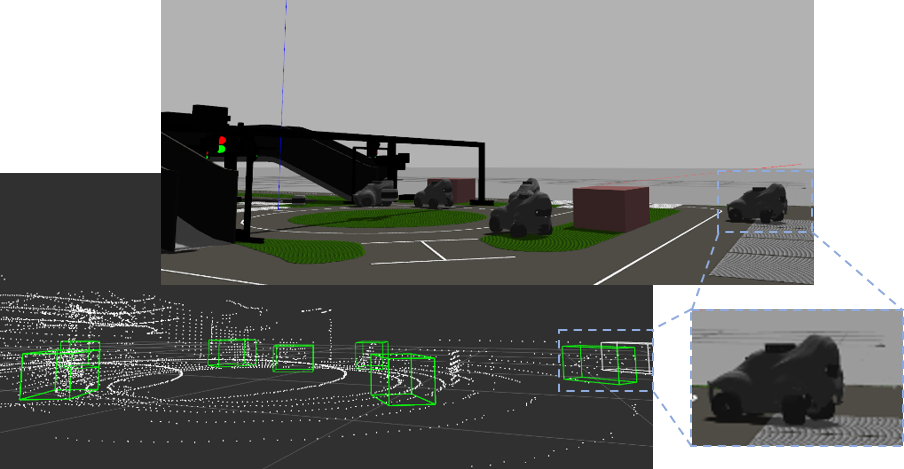}
        \caption{Gazebo simulation scenario and 3D detection results}
        \label{3D detection results}
    \end{figure}

    \begin{figure*}
        \includegraphics[width=\linewidth]{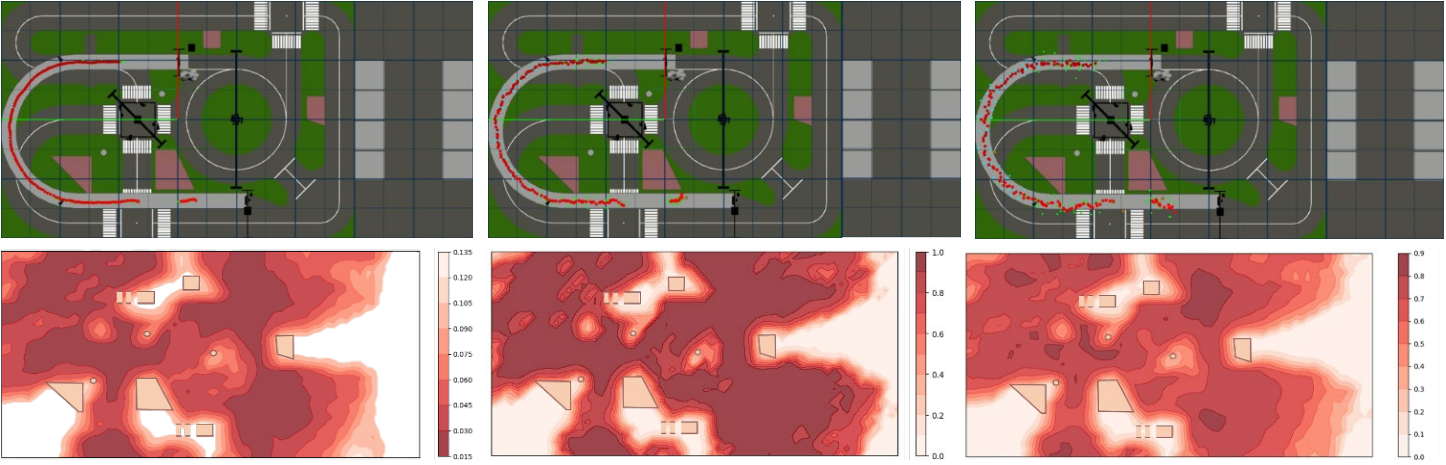}
        \caption{(Top) KF with different Gaussian noises of variance=0, variance=0.05, variance=0.1 respectively. Points in different color: green points---observation of single sensor, cyan points---observation of multiple sensors and red points---output of the Kalman filter; (Bottom) Lidar-camera fusion detection upon the whole scene, metrics from left to right: pose error, AOS, and IOU of BEV.}
        \label{exps}
    \end{figure*}

Figure \ref{3D detection results} shows the detection results of one frame, where the green boxes represent the results after successful matching and merging, while the white boxes represent the failures ones. Note the only one car in the far right corner, with the misdetected result successfully removed through multi-sensor detection.

\section{MULTI-AGENT COLLABORATIVE SCHEDULING AND DECISION}

\subsection{Agent Division}

In the study of \cite{c23}, the author proposed to switch between V2I mode and V2V mode as the traffic flows change to find the lowest delay and the most reliable decisions. This paper further extends its V2I mode, dividing the complex traffic scene into different subjects, each of which contains a roadside unit (RSU) consists of a sensor detection system and an edge computing platform.

\begin{figure}[thpb]
        \centering
        \includegraphics[scale=0.45]{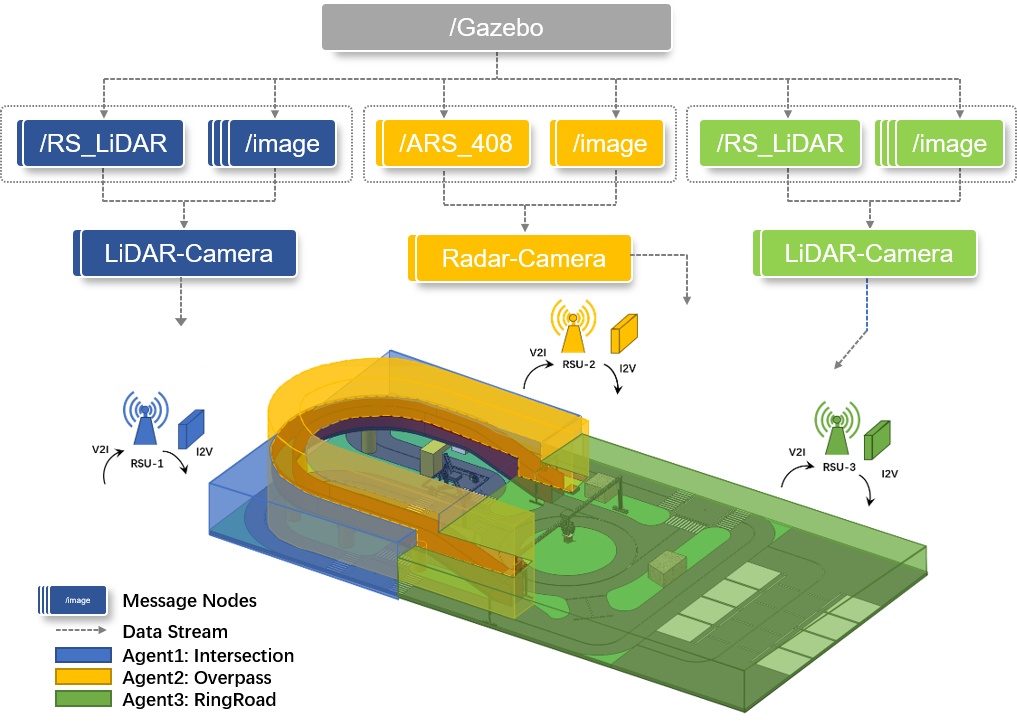}
        \caption{Data Stream and Three Agents.}
        \label{Data Stream and Three Agents.}
    \end{figure}

The bases for dividing the main body are the traffic function, spatial hierarchy, or the distance. So we introduce the agent division to handle the long range of distance and functional differences. See in Fig.\ref{Data Stream and Three Agents.}, the agents consist of three parts: agent1---the intersection, agent2---the overpass ramp, and agent3---the ring road. The three subjects complete the perception, fusion, and scheduling within their respective areas. In the decision stage, the information is exchanged and merged on the cloud platform to make a collaborative decision. The overall information transmission nodes and topic framework of the system are also shown in Fig.\ref{Data Stream and Three Agents.}.

\subsection{Cooperative Scheduling}

Given the agents covering the whole field, we link it to the perception and fusion system. For simplicity, we number and label all the roads and areas where vehicles may occur the traffic jam. From the information offered by perception fusion system, agents could directly build a vehicle density and speed distribution map, hence then transmit the messages that point out where to avoid the peak traffic or else. For our follow-up experiments, we make the vehicles led by agents to optimize the route at any time.

\section{EXPERIMENTS} \label{Experiments}

\subsection{Implementation Details}

\textbf{Datasets.} We build our image detection datasets through blender and gazebo. The collections consist of 13985 images all of which are annotated as VOC format. Train, test and validate split account for 6300, 4285 and 3400 respectively. These images are given by various perspectives and distances, untruncated and partially truncated. And it covers almost the states of the vehicles driving under all the roads.

    \begin{figure*}
    \includegraphics[width=\linewidth]{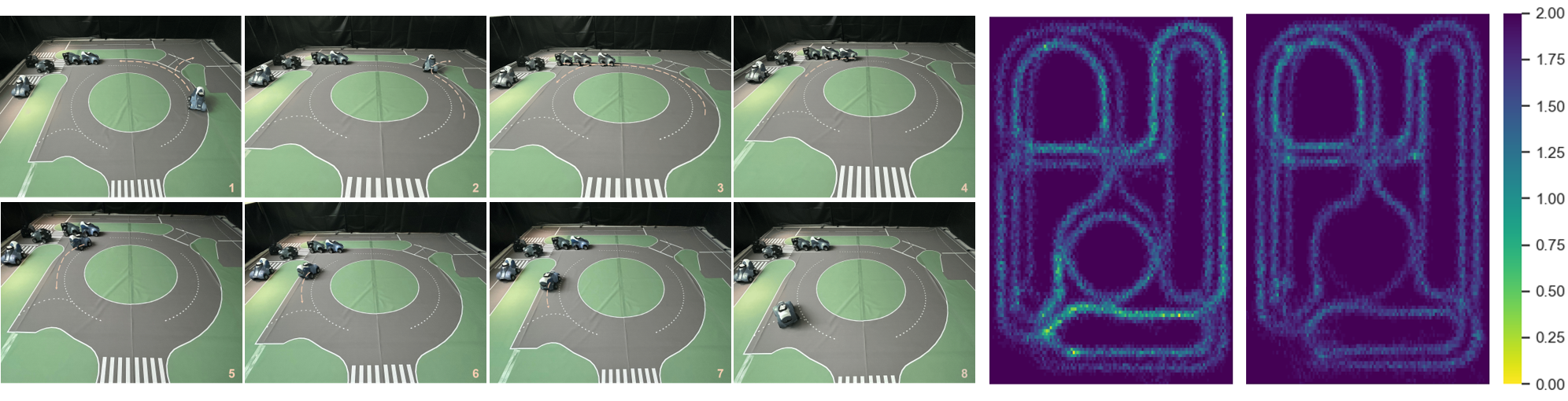}
    \caption{(Left) Vehicles Improved by Roadside Unit; (Right) Vehicles Speed Map: Randomly driving (left) and Scheduleing driving (right); Lighter color shows lower speed while darker color shows faster speed. Large number of vehicles queuing at the intersection mainly results in low speed.}
    \label{real and heatmap}
    \end{figure*}

For 3D point cloud detection, we shot the frames of them as they run in the environment similar to image shots and annotated as KITTI \cite{c27} format using LabelCloud toolbox \cite{c28}. Finally we used 1750 frames of the simulation scene and 1250, 550 of them for training and test. respectively. And data augmentations \cite{c29} are applied to improve the model generalization performance.

\textbf{Model.} We adapted the configurations from PointPillars \cite{c5} to train our 3D detector. For our own datasets, we set the detection range to $[-12m, 12m]$ for X and Y axes, and $[-1m, 4m]$ for Z axis. We use $(0.015m, 0.015m)$ as the pillar size for experiments. For other default settings, the readers could refer to the OpenPCDet toolbox \cite{c30} used in this work.

\textbf{Sim-to-Real.} To valid the comprehensive function of our system, we performed successful sim-to-real transfer and built a real $8m\times 10m$ sand table consistent with the simulation. We use PC as the MEC and communicate with AWS Deepracers via AIoT technology.

\subsection{Muti-type and Muti-group Sensor Detection}
In the simulation, gaussian noises are added to evaluate the KF. The performance of KF is shown in Fig.\ref{exps} (top), where points in different color: green points---observation of single sensor, cyan points---observation of multiple sensors and red points---output of the Kalman filter, showing the KF denoises effectively. In this case, the fusion can balance the noise influence and filter the false detection that may occur irregularly due to the hardware. Once any of detectors fails, the other can ensure the overall detection effectiveness.

Fig.\ref{exps} (bottom) shows the performance of lidar-camera detection of $2.9\times10^4$ consecutive frames. Fig.\ref{exps} (top) has the bird view of the overall scene, which acts as a geographical reference. The thermal map on the left shows the average pose error represented by Euclidean distance. We take the odometry information released by the vehicle itself as the ground truth. It can be seen that the error of most areas is within $0.045m$.The map in the middle shows the average orientation similarity (AOS) of detections, using the cosine distance of the azimuth angle difference (Eq.\ref{eq: AOS}), and most parts excess $0.9$. Similarly, the BEV IOUs of boxes are mostly greater than $0.7$. 
\begin{equation} \label{eq: AOS}
    AOS = \frac{1}{|D|}\Sigma_{s\in D}\frac{1+cos(2\Delta^\theta_s)}{2}
\end{equation}

\subsection{Scheduling System}

To apply the detection system to the actual traffic flow and perform the downstream tasks, we attached a lightweight scheduling system with three agents covering the whole scene. We use AIoT to associate real vehicles with simulated environments and conduct experiments simultaneously, thus through the real vehicles we can feel more intuitively and preliminarily verify the feasibility of our scheme in real world. Fig.\ref{real and heatmap} (Left) introduces one scenario, with 6 vehicles heading the same intersection.

    \begin{figure}[thpb]
        \centering
        \includegraphics[scale=0.45]{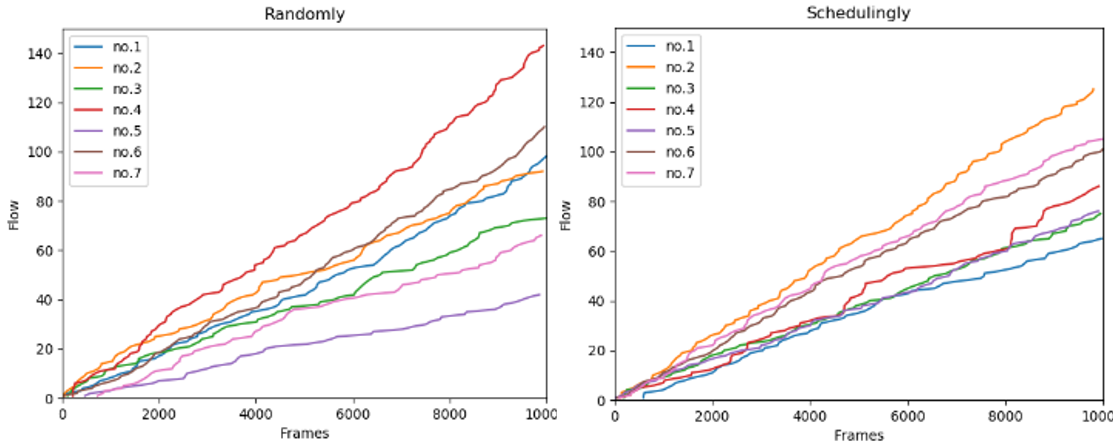}
        \caption{Trafic Flow of 7 intersections: From random(left) to scheduling(right) - reduce the heavy load while improve the use of roads with light load.}
        \label{Trafic Flow of 7 intersections}
    \end{figure}

We focus on the action of one vehicle, located on the right of frame 1. The critical points are the frame 2 and 4, where the vehicle faced with the possibility of joining the traffic jam. With the perception system offering the global information and the scheduling system offering route plans, most vehicles could always get what they want---to avoid congestion while reach the least time cost. Here we only conduct a simple one---to showcase the functionality and work process of our proposed system, under the limited conditions.

In the simulation, we conducted automatic driving of 10 vehicles randomly and schedulingly. Fig.\ref{real and heatmap} (Right) shows the mean speed of all vehicles over $2.9\times10^4$ frames, where a great reduction on traffic load can be seen.

As depicted in Fig.\ref{Trafic Flow of 7 intersections}, the other proof is the flows of different road intersections over time. Under the help of global detection and  scheduling system, the utilization rate of different roads is optimized, with the increased total flow ($624$ for random and $668$ for scheduling, see Table \ref{Traffic flow before and after scheduling}). 

\begin{table}[h!]
    \centering
    \caption{Traffic flow before(B.) and after(A.) scheduling}
    \label{Traffic flow before and after scheduling}
    \begin{tabular}{ c|c c c c c c c|c } 
        \hline
        \rule{0pt}{7pt}
        & No.1 & No.2 & No.3 & No.4 & No.5 & No.6 & No.7 & total \\
        \hline
        \rule{0pt}{7pt}
        B. & 100 & 92 & 72 & 144 & 42 & 110 & 63 & 624\\
         A. & 73 & 122 & 87 & $\textcolor{green}{94}\downarrow$ & $\textcolor{green}{88}\uparrow$ & 100 & 104 & $\textcolor{green}{668}\uparrow$ \\
        \hline
    \end{tabular}
\end{table}

Finally, we achieve a success of the whole system, get $FPS \sim10$---the biggest loss lies in time synchronization among the three sensors, yet we reach the real-time run on RMMDet system.

\section{CONCLUSIONS AND DISCUSSIONS}

In this paper, we present RMMDet, a road-side multitype and multigroup sensor detection system for autonomous driving, for more intelligent vehicle-road collaboration. RMMDet uses a virtual simulation counterpart based on ROS, implementing three types sensors detection and lidar-camera, radar-camera fusion detection in the complex traffic scene. Real-time multimodal data streams are captured, processed, fed into model and generate the desired results to construct an end-to-end efficient roadside system. For the detection algorithms, we initially choose YOLOv4 and pointpillars, which could be replaced by any other methods and it's convenient for further research and experiments. And we link a lightweight scheduling system to showcase the functionality.

In general, RMMDet presents an early attempt to build a comprehensive roadside system virtually and actually for autonomous driving. For further improvements, we plan to take more advanced algorithms into account and transfer the result-level fusion method to more learnable methods. And conduct more complex experiments under real conditions with more resources. This is a long-line work and we will keep on it.

\addtolength{\textheight}{-12cm}   % This command serves to balance the column lengths


% \section*{ACKNOWLEDGMENT}

% The preferred spelling of the word ÒacknowledgmentÓ in America is without an ÒeÓ after the ÒgÓ. Avoid the stilted expression, ÒOne of us (R. B. G.) thanks . . .Ó  Instead, try ÒR. B. G. thanksÓ. Put sponsor acknowledgments in the unnumbered footnote on the first page.

% References are important to the reader; therefore, each citation must be complete and correct. If at all possible, references should be commonly available publications.

\end{document}